\documentclass[10pt,twocolumn,letterpaper]{article}

\usepackage{iccv}
\usepackage{times}
\usepackage{epsfig}
\usepackage{graphicx}
\usepackage{amsmath}
\usepackage{amssymb}

\usepackage{caption}
\usepackage{graphicx}
\usepackage{booktabs}
\usepackage{pifont}
\usepackage{multirow}
\usepackage{bm}
\usepackage[table,xcdraw]{xcolor}

\usepackage{dsfont}
\usepackage{graphics}
\usepackage[export]{adjustbox}
\usepackage{diagbox}

\usepackage[breaklinks=true,bookmarks=false]{hyperref}

\iccvfinalcopy % *** Uncomment this line for the final submission

\def\iccvPaperID{****} % *** Enter the ICCV Paper ID here

\ificcvfinal\fi

\definecolor{pltblue}{RGB}{174, 199, 232}
\definecolor{pltorange}{RGB}{255, 229, 204}
\definecolor{pltgreen}{RGB}{204, 229, 204}
\definecolor{pltred}{RGB}{229, 204, 204}
\definecolor{pltpurple}{RGB}{239, 218, 230}

\definecolor{tabblue}{HTML}{1f77b4}
\definecolor{taborange}{HTML}{ff7f0e}
\definecolor{tabgreen}{HTML}{2ca02c}
\definecolor{tabred}{HTML}{d62728}
\definecolor{tabpurple}{HTML}{9467bd}
\definecolor{tabpink}{HTML}{ff0080}

\definecolor{cblue}{RGB}{173, 201, 233}
\definecolor{clblue}{RGB}{222, 234, 246}
\definecolor{corange}{RGB}{255, 152, 67}
\definecolor{lorgange}{RGB}{255, 221, 149}

\definecolor{ovcdred}{RGB}{227, 115, 139}
\definecolor{ovcdgreen}{RGB}{123, 196, 197}
\definecolor{ovcdblue}{RGB}{140, 165, 234}

\definecolor{ovcdgray}{RGB}{150, 150, 150}
\definecolor{ovcd_building}{RGB}{255, 171, 124}
\definecolor{ovcd_water}{RGB}{151, 210, 255}
\definecolor{ovcd_playground}{RGB}{239, 171, 178}

\newcommand{\MCI}{\textcolor{ovcdred}{\textbf{M}}{-}\textcolor{ovcdgreen}{\textbf{C}}{-}\textcolor{ovcdblue}{\textbf{I}}}
\newcommand{\IMC}{\textcolor{ovcdblue}{\textbf{I}}{-}\textcolor{ovcdred}{\textbf{M}}{-}\textcolor{ovcdgreen}{\textbf{C}}}

\begin{document}

%%%%%%%%% TITLE
\title{\textit{Dynamic\textcolor{tabblue}{Earth}}: How Far are We from Open-Vocabulary Change Detection?}

\makeatletter
\def\thanks#1{\protected@xdef\@thanks{\@thanks
        \protect\footnotetext{#1}}}
\makeatother

\author{Kaiyu Li$^1$, Xiangyong Cao$^{1\dag}$, Yupeng Deng$^2$, Chao Pang$^3$, Zepeng Xin$^1$, Deyu Meng$^1$, Zhi Wang$^1$ \\
$^1$Xi’an Jiaotong University \; $^2$Chinese Academy of Sciences \; $^3$Wuhan University\\
{\tt\small likyoo.ai@gmail.com, caoxiangyong@mail.xjtu.edu.cn, dengyp@aircas.ac.cn} \\
{\tt\small pangchao@whu.edu.cn, 37xinzepeng@stu.xjtu.edu.cn, dymeng@mail.xjtu.edu.cn, zhiwang@xjtu.edu.cn} \\
Project: \textcolor{tabred}{\url{https://likyoo.github.io/DynamicEarth}}
}

% \maketitle

\twocolumn[{%
\renewcommand\twocolumn[1][]{#1}%
\maketitle
\vspace{-10mm}
\begin{center}
  \centering
  \captionsetup{type=figure}
  \includegraphics[width=0.95\linewidth]{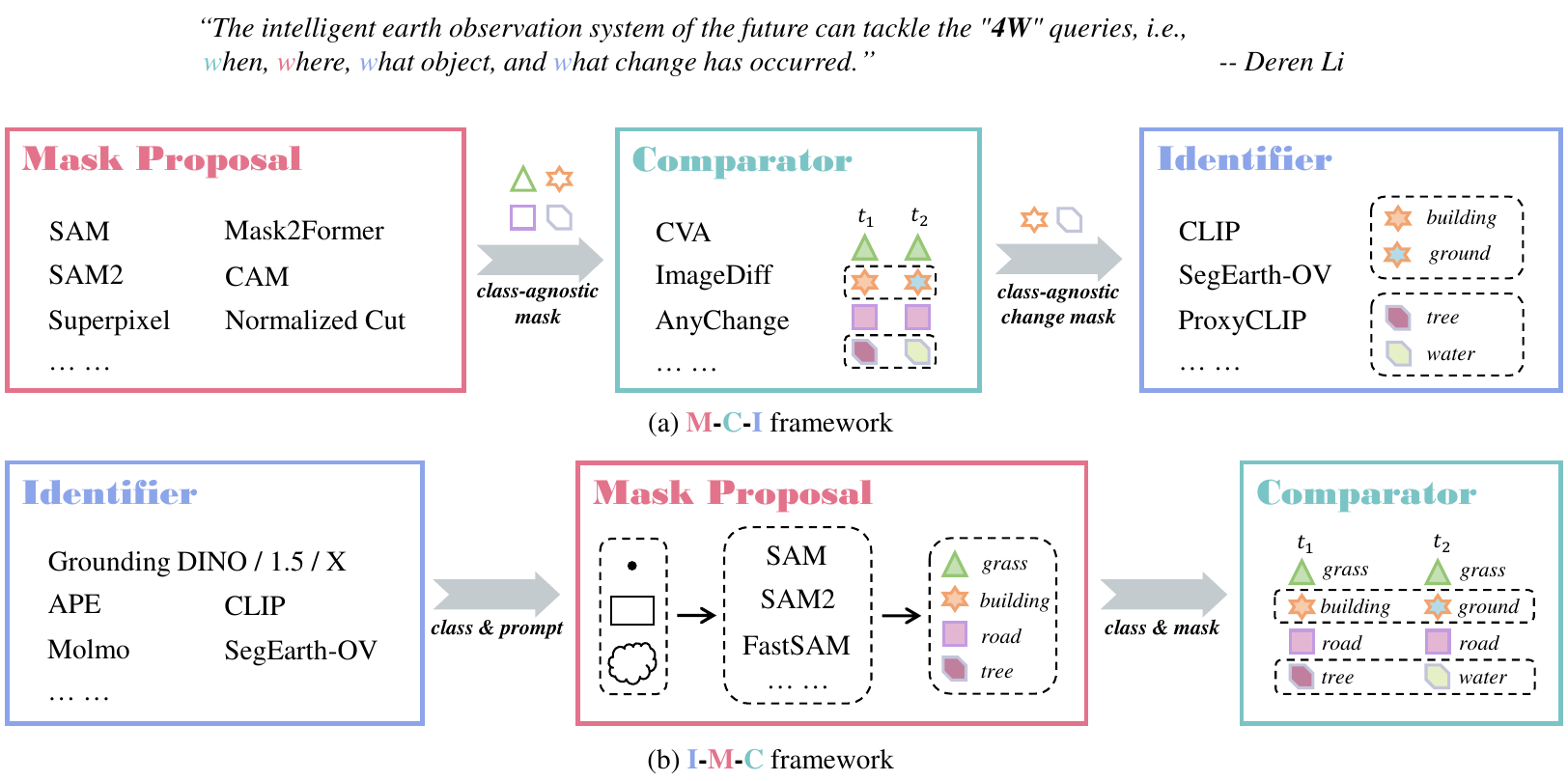}
  \captionof{figure}{The two OVCD frameworks proposed in this paper. (a) \MCI: discover all class-agnostic masks, determine if the mask region has changed, and identify the change class. (b) \IMC: identify all targets of interest, convert to mask format, and compare if the target has changed.}
  \label{fig:fig_title}
\end{center}%
}]

\def\thefootnote{}\footnotetext[1]{$^\dag$ Corresponding author.}
\def\thefootnote{\arabic{footnote}}

% Remove page # from the first page of camera-ready.
\ificcvfinal\thispagestyle{empty}\fi

%%%%%%%%% ABSTRACT
\begin{abstract}
   Monitoring Earth's evolving land covers requires methods capable of detecting changes across a wide range of categories and contexts. Existing change detection methods are hindered by their dependency on predefined classes, reducing their effectiveness in open-world applications. To address this issue, we introduce open-vocabulary change detection (OVCD), a novel task that bridges vision and language to detect changes across any category. Considering the lack of high-quality data and annotation, we propose two training-free frameworks, \MCI~and \IMC, which leverage and integrate off-the-shelf foundation models for the OVCD task. The insight behind the \MCI~framework is to discover all potential changes and then classify these changes, while the insight of \IMC~framework is to identify all targets of interest and then determine whether their states have changed. Based on these two frameworks, we instantiate to obtain several methods, \textit{e.g.}, \textcolor{ovcdred}{SAM}{-}\textcolor{ovcdgreen}{DINOv2}{-}\textcolor{ovcdblue}{SegEarth-OV}, \textcolor{ovcdblue}{Grounding-DINO}{-}\textcolor{ovcdred}{SAM2}{-}\textcolor{ovcdgreen}{DINO}, \textit{etc}. Extensive evaluations on 5 benchmark datasets demonstrate the superior generalization and robustness of our OVCD methods over existing supervised and unsupervised methods. To support continued exploration, we release \textit{Dynamic\textcolor{tabblue}{Earth}}, a dedicated codebase designed to advance research and application of OVCD.
\end{abstract}

%%%%%%%%% BODY TEXT
\section{Introduction}

The Earth is a dynamic system in a state of perpetual evolution. Observing this vibrant planet allows us to deepen our understanding of intricate phenomena, including human activity, geographic evolution, and climate change. This process requires the continuous monitoring of land use and land cover types and provides insights into where changes have occurred and the particulars of those changes. The most pertinent task in this context is change detection, which specifically involves analyzing bi-temporal or multi-temporal satellite and aerial images to determine what changes are occurring where.

As a higher-level task beyond segmentation and detection, change detection technology has made significant strides with the support of basic vision techniques. Typically, under the supervised learning paradigm, some methods take bi-temporal images into neural networks and predict their change masks or bounding boxes~\cite{fang2023changer, chen2024rsmamba, deng2022feature, Chen2024ObjFormer}. However, the generalization of the models developed within this paradigm is limited. Even for the same category (\eg building), these models can be challenging to apply directly to images captured with different cameras, ground sample distances (GSDs), and other variations. Recently, with the emergence of foundation models, some advanced methods have achieved improved generalization in supervised~\cite{li2024new, chen2024time}, semi-supervised~\cite{li2024semicd, yang2024unimatch}, and unsupervised~\cite{tan2023segment, zheng2024segment} change detection by leveraging the general knowledge of these foundation models. In particular, AnyChange~\cite{zheng2024segment} employs the mask proposal and feature matching of the segment anything model (SAM) \cite{kirillov2023segment} to construct a universal unsupervised change detection model without any post-training. However, it is limited in that AnyChange can only address the binary change detection problem, as it generates class-agnostic change masks. In other words, while AnyChange effectively identifies ``where'' changes occur, it does not provide insights into ``what'' those changes are. Motivated by this, in this paper, we introduce a new task aimed at discovering changes in any category of interest, which we term \textbf{Open-Vocabulary Change Detection (OVCD)}. We also try to explore how far current methodologies are from achieving OVCD.

Different from open-vocabulary segmentation or detection of single-temporal images \cite{zhu2024survey}, OVCD involves the identification and comparison of bi- or multi-temporal images. In addition, \cite{li2024semicd} and \cite{zheng2024segment} emphasize that instance-level comparisons are generally superior to simple pixel-level comparisons in change detection, effectively mitigating pixel-level pseudo changes that may arise from changes in lighting, season, viewpoint, \etc. Consequently, the proposal of instance-level objects/masks should also be regarded as an essential element of OVCD. Accordingly, we advocate that a complete OVCD framework should consist of three components: an \textcolor{ovcdblue}{\textbf{identifier}}, a \textcolor{ovcdgreen}{\textbf{comparator}}, and the \textcolor{ovcdred}{\textbf{proposal of instance-level objects/masks}}. On the other hand, considering the severe lack of high-quality annotations for change detection, we do not suggest training a specialized vision-language model (VLM) for change detection from scratch, but rather reusing off-the-shelf general VLMs.

Based on the above, in this paper, we propose two training-free universal frameworks for OVCD, \MCI~and \IMC, as shown in Figure \ref{fig:fig_title}. \textbf{(1)} \MCI~framework first generates class-agnostic masks using mask proposal methods, \eg, SAM~\cite{kirillov2023segment}, Mask2Former~\cite{cheng2022masked}, \etc. It then compares the region corresponding to each mask in the bi-temporal images or features to determine whether a change has occurred. If a change is detected, the mask region is fed into the final open-vocabulary classifier (\eg, CLIP~\cite{radford2021learning}, SegEarth-OV \cite{li2024segearth}) to identify the change category. \textbf{(2)} \IMC~framework is inspired by the post-classification comparison (PCC) method~\cite{howarth1981procedures}. It begins by using perceptual foundation models, \eg, grounding DINO~\cite{liu2025grounding}, APE~\cite{shen2024aligning}, or Molmo~\cite{deitke2024molmo}, which utilize textual input to guide the identification of objects in the format of bounding box, coarse mask, or point. These are then used as visual prompts in the mask proposal method to yield the fine-grained masks. Finally, the masks corresponding to the same positions in the bi-temporal images are compared to determine whether any changes have occurred.

To summarize, the contributions of this work include:

\textbf{(i)} A new task, OVCD, is introduced, which unlocks language-guided change detection and allows detection of changes in any category.

\textbf{(ii)} Two training-free universal frameworks for OVCD are proposed which fully reuse the off-the-shelf foundation models. 

\textbf{(iii)} Extensive evaluations on diverse datasets highlight the generalization and robustness of our method, significantly surpassing existing unsupervised and supervised methods.

\section{Related work}

\begin{figure*}[t]
  \centering
   \includegraphics[width=1.\linewidth]{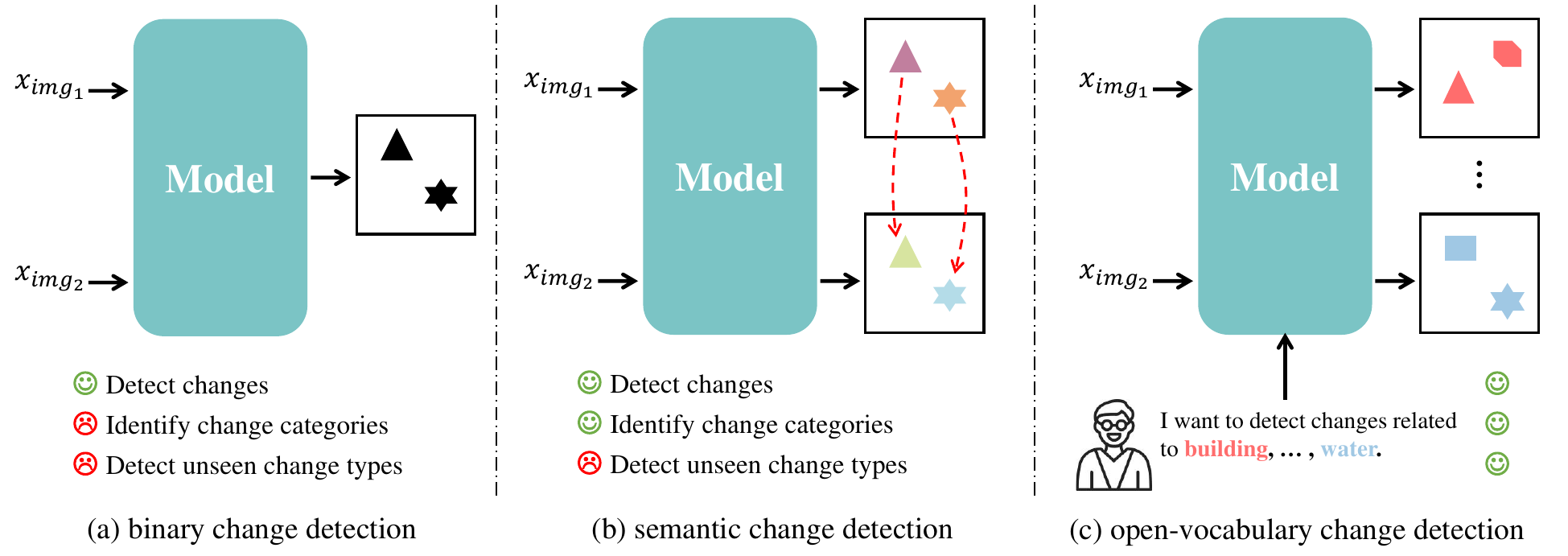}
   \caption{Different change detection tasks: (a) Binary change detection aims at discovering all (interested) changes and generating a binary mask; (b) Semantic change detection further identifies the category of changes. However, both can only be trained and evaluated on data with predefined categories; (c) Our proposed OVCD can detect changes in any category according to the user's requirements.}
   \label{fig:ovcd}
\end{figure*}

\noindent
\textbf{Vision-language model.}
VLMs aim to bridge the gap between visual and textual modalities, enabling systems to understand and generate multi-modal content. Pioneering works such as CLIP~\cite{radford2021learning} and SigLIP~\cite{zhai2023sigmoid} introduced methods to align visual and linguistic representations through contrastive learning. Inspired by the large language model (LLM)~\cite{touvron2023llama, team2024gemma}, some methods~\cite{liu2024improved, liu2024visual, xue2024xgen} focus on both comprehension and generation, answering diverse tasks in the form of multi-modal chats, including object types, object counts, object locations, \etc. More recently, under this paradigm, some efforts towards ``one-for-all'' perceptual models have emerged. Grounding DINO~\cite{liu2025grounding, ren2024grounding} marries DINO~\cite{zhang2022dino} with grounded pre-training for open-set object detection. Then, the subsequent APE~\cite{shen2024aligning} and DINO-X~\cite{ren2024dino} allow fine-grained perception \eg segmentation or keypoints. Further, the advanced Florence-2~\cite{xiao2024florence} takes textual prompts as task instructions and generates desirable results in text form, including captioning, object detection, grounding and segmentation.

% These models have demonstrated remarkable generalization capabilities across tasks such as zero-shot image classification and visual question answering~\cite{liu2024improved, liu2024visual}. More recently, Grounding DINO~\cite{} marries DINO~\cite{} with grounded pre-training for open-set object detection.

\noindent \textbf{Segment anything model.}
SAM~\cite{kirillov2023segment} pioneers a new segmentation paradigm, utilizing prompt-based learning to enable segmentation using points, boxes, or masks as inputs. Unlike traditional methods that rely on domain-specific training datasets, SAM leverages vast amounts of pre-trained data to achieve high generalizability. Based on this, HQ-SAM~\cite{ke2024segment} introduces HQ-Token for high-quality mask prediction. Recent advancements include models \eg, FastSAM~\cite{zhao2023fast}, MobileSAM~\cite{zhang2023faster}, EfficientViT-SAM~\cite{zhang2024efficientvit}, and TinySAM~\cite{shu2023tinysam}, which focus on improving efficiency, scalability, and real-time performance. Most recently, SAM~2~\cite{ravi2024sam} has been proposed to further extend to video equipped with memory attention mechanism, while keeping the original ability to segment everything in images. In OVCD, SAMs can be used either as an initial step to propose all candidate regions or as a post-refiner of the identification results.

% These lightweight and optimized versions have extended the applicability of SAMs to resource-constrained and real-time scenarios.

\noindent
\textbf{Open-vocabulary semantic segmentation.}
OVSS extends semantic segmentation to recognize and segment unseen categories at inference time. Most OVSS methods build on top of VLMs. Early OVSS methods are inspired by vision-language contrastive learning and try to train the CLIP variant with pixel-level perception~\cite{li2022language, xu2022groupvit, xu2023side, liang2023open, cho2024cat}. Due to the domain difference between remote sensing images and natural images, Cao \etal~\cite{cao2024open} proposed a CLIP fine-tuning based method and trained it on some remote sensing segmentation data. Similarly, Ye \etal~\cite{ye2025GSNet} used more remote sensing data (consisting of public datasets) to train both CLIP and a specialist backbone. Different from the above, Li \etal~\cite{li2024segearth} found that feature resolution is the key factor constraining OVSS of remote sensing images, and proposed a training-free OVSS method, SegEarth-OV, which even outperforms the training-based method~\cite{lan2025proxyclip, wang2025sclip, lan2025clearclip, zhang2024corrclip}. In this paper, we believe that identification is an important component of OVCD, and thus to some extent, a feasible OVSS method is the prerequisite for OVCD.

\noindent
\textbf{Binary change detection.}
Binary change detection involves identifying regions of change between two temporally separated images. Traditional methods \eg, change vector analysis (CVA)~\cite{bovolo2006theoretical} relied on pixel-wise comparisons, while more recent methods leverage deep learning for enhanced feature extraction and robustness. Siamese networks~\cite{bromley1993signature} and attention-based models~\cite{vaswani2017attention, fang2021snunet, fang2023changer} have gained popularity for their ability to learn discriminative change representations. In the foundation model era, BAN~\cite{li2024new} introduces VLMs to binary change detection and provides a supervised parameter efficient fine-tuning solution. Subsequently, Li \etal~\cite{li2024semicd} used the pseudo labels generated by VLMs as additional supervised signals to improve the performance of semi-supervised change detection. Zheng \etal~\cite{zheng2024segment} proposed AnyChange for unsupervised binary change detection, which yields high-recall class-agnostic change masks. However, in practical cases, the extraction of changes of interest is essential, and our proposed \MCI~framework can be regarded as an upgrade and generalization of AnyChange, \ie, it can segment any change of interest.

\noindent
\textbf{Semantic change detection.}
Semantic change detection builds on binary change detection by not only identifying changes but also classifying them into semantic categories. As mentioned above, compared to binary change detection, semantic change detection is more in line with practical applications. Users often only need to detect changes in some specific categories, \eg, building change detection for urban expansion analysis~\cite{ji2018fully, chen2020spatial, pang2023detecting}, cropland change detection for agricultural protection~\cite{liu2022cnn, sun2024identifying}, landslide change detection for disaster monitoring~\cite{zhang2023cross}, \etc. Typical semantic change detection methods follow a triple decoder architecture, \ie, one difference branch for binary change detection and two semantic branches for bi-temporal class discrimination~\cite{yang2020semantic, zheng2022changemask, deng2022feature, deng2023tchange}. In this paper, we simplify the multi-class semantic change detection to single-class semantic change detection, which achieves the same objective while avoiding the complexity of the model structure, and allows the extraction of changes in any class using only off-the-shelf single-temporal models.

% TODO
% SCM

\section{Problem Definition}

OVCD aims to localize and identify changes between two temporally separated images $x_{{img}_1}$ and $x_{{img}_2}$ of the same scene, where the categories of changes are not predefined and can be described by arbitrary textual or semantic labels $x_{text}$. OVCD extends traditional change detection by introducing the ability to generalize beyond a fixed set of predefined change classes, enabling the detection and interpretation of novel changes using textual guidance or contextual understanding, as shown in Figure~\ref{fig:ovcd}. The task shares a similar formulation with the OVSS task~\cite{xu2023side, li2024segearth}, but is far more challenging. The key challenge lies in:
\begin{itemize}
    \item \textbf{Bi- or multi-temporal image input.} Beyond OVSS, OVCD involves comparisons between image pairs. This introduces several questions, \eg, when should comparisons be made, before or after identification? What level of comparison should be performed, pixel level or instance level? How to mitigate error accumulation due to multiple steps? And so on.
    \item \textbf{Cross-domain from natural to remote sensing images.} Remote sensing images are generally acquired by sensors from satellites or aircraft, and they present a bird's eye view, as different from natural images which mostly present a horizontal view. These two views bring completely different surface features when facing the same object, \eg, for a building, it may be characterized by a window in the horizontal view and a roof in the overhead view. This difference leads to the fact that most of the models trained in natural images cannot be directly applied to remote sensing images.
    \item \textbf{Other challenges.} There are also some challenges due to the characteristics of remote sensing images, \eg, OVCD of non-RGB images (\ie, multispectral, hyperspectral, SAR images, \etc.), object scale spanning and small object issues in overhead views.
\end{itemize}

\section{Method}

%  Most recently, the emergence of some VLMs has led to a turnaround, \eg, Li \etal~\cite{li2024semicd} found that APE~\cite{shen2024aligning} exhibit generalization capabilities in simple remote sensing images.

Following the experience of OVSS, there are two basic paths to achieve OVCD: training-based method and training-free method. Training-based methods generally train on some basic classes and then generalize to novel classes by leveraging the cross-modal and zero-shot capabilities of VLMs. However, the current volume of change detection data, especially semantic change detection data, is limited and dispersed under various GSDs~\cite{peng2025deep}. In addition, the image quality and annotation quality are worrisome~\cite{tian2020hi, tian2022large}. These factors make it difficult to implement a complete training-based OVCD pipeline. A contemporaneous work, Semantic-CD~\cite{zhu2025semantic}, also attempted an open-vocabulary setup, but was limited by the data in that it was only trained and evaluated on the same dataset, which undermines the assertion of true open-vocabulary capability. On the other hand, some recent studies have demonstrated that foundation models trained on web data or natural image data also exhibit generalization capabilities in remote sensing scenarios~\cite{shen2024aligning, li2024semicd, li2024segearth, zheng2024segment}. For instance, SegEarth-OV~\cite{li2024segearth} indicates that standard CLIP~\cite{radford2021learning} can be directly used for semantic segmentation of remote sensing images without any post-training.

Therefore, in this paper, we focus on \textbf{\textit{how to construct training-free OVSS methods using off-the-shelf foundation models}}. Based on the definitions of change detection and OVCD, we find two strategies to achieve ``segment any change of interest'': (1) discover all potential changes, and then classify these changes; (2) identify all targets of interest, and then determine whether their states have changed between two images. Therefore, a \textcolor{ovcdgreen}{\textbf{comparator}} and an open-set \textcolor{ovcdblue}{\textbf{identifier}} are necessary to implement these two strategies. In addition, in practice, the \textcolor{ovcdred}{\textbf{mask proposal}} is also needed. For the first strategy, the mask proposal can mitigate pixel-level pseudo changes that may arise from changes in lighting, season, viewpoint, \etc. For the second strategy, the mask proposal can convert the bounding boxes and points in the identification results into uniform masks, aligning more effectively with the task requirements. According to the above two strategies, we propose two training-free OVCD frameworks, \MCI~and \IMC, as shown in Figure~\ref{fig:fig_title}.

\subsection{\MCI}

The \MCI~framework pipeline begins with discovering all class-agnostic masks. Following this, it compares whether the bi-temporal features corresponding to these masks have changed. Finally, it filters the changes of interest. Based on this, the \MCI~framework can be decomposed into three sequential components: \textcolor{ovcdred}{\textbf{Mask Proposal}}, \textcolor{ovcdgreen}{\textbf{Comparator}}, and \textcolor{ovcdblue}{\textbf{Identifier}}.

\noindent
\textcolor{ovcdred}{\textbf{Mask Proposal}} component in the \MCI~framework is designed to divide the image into several partitions $\mathcal{M}_{t}=\{\mathbf{m}_{t, i}\}_{i \in[1,2, \ldots, N_{t}]}$, where $t$ indicates the temporal index, and expects each partition to contain an instance or a single category of pixels. In fact, before the deep learning era, some traditional image segmentation algorithms had this capability, \eg, Normalized Cuts~\cite{shi2000normalized}, SuperPixel~\cite{achanta2012slic}, Ncuts~\cite{ren2003learning}, \etc. These methods only utilize texture properties to segment the image and thus have limited performance on complex and low-resolution images. A better alternative is SAMs, which are trained on a large number of high-quality labeled images (\eg, SAM~\cite{kirillov2023segment} uses 11M images with 1.1B masks, SAM~2~\cite{ravi2024sam} uses 1.4K videos with 16K masklets), and thus offer strong generalization and fine segmentation~\cite{chen2024rsprompter, ren2024segment}. In this framework, SAMs are required to enable the automatic mask generation mode to generate all candidate masks. Since the changed objects may appear in either temporal image, we concatenate the bi-temporal candidate mask sets as an overall candidate mask set. Considering that the bi-temporal images are captured from the same region and contain numerous unchanged objects, we use non-maximum suppression (NMS) to remove duplicate masks. Specifically, to enhance efficiency, we use the outer bounding boxes and the IoU predictions of candidate masks as inputs to the NMS. Consequently, the final output of the Mask Proposal is denoted as $\mathcal{M}=\textit{NMS}(\mathcal{M}_1 \cup \mathcal{M}_2)$.

\noindent
\textcolor{ovcdgreen}{\textbf{Comparator}} is used to discriminate whether the mask region has changed or not, and its purpose is the same as the binary change detection task. Therefore, some traditional change detection methods can be used as comparators, \eg, change vector analysis (CVA)~\cite{malila1980change}, ratioing method~\cite{lu2004change}, differencing method~\cite{mahmoudzadeh2007digital}, \etc. In our implementation, following~\cite{zheng2024segment}, we use a latent matching method that use negative cosine similarity as the change score for bi-temporal latent features in the mask region (higher score indicates more significant change). But unlike~\cite{zheng2024segment}, we suggest to use DINO/DINOv2~\cite{zhang2022dino, oquab2023dinov2} to extract features, considering the natural advantage of DINOs, \ie, they are trained using contrastive learning, where the feature distances of the same objects are pulled closer together, and vice versa. Specifically, the change score can be formulated as:

\begin{equation}
\begin{aligned}
\mathcal{D}\left(\mathbf{m}\right)=-\frac{\mathbf{z_{1}}[\mathbf{m}]}{\left\|\mathbf{z_{1}}[\mathbf{m}]\right\|_{2}} \cdot \frac{\mathbf{z_{2}}[\mathbf{m}]}{\left\|\mathbf{z_{2}}[\mathbf{m}]\right\|_{2}},
  \label{eq:change_score}
\end{aligned}
\end{equation}
where the vector $\mathbf{z_{t}}[\mathbf{m}]$ indicates the average of the feature maps from $x_{{img}_t}$ extracted using DINOs at the indexes corresponding to the mask $\mathbf{m}$. $\mathbf{m} \in \mathcal{M}$ and $t \in \{1, 2\}$. The masks with change scores higher than the threshold $\beta$ will be discriminated as change masks and the rest are dropped.

\noindent
\textcolor{ovcdblue}{\textbf{Identifier}} is used to filter specific categories of changes from all change masks. Since the categories of interest to the user may be varied, it is essential to employ VLMs with open-set identification capabilities, \eg, CLIP~\cite{radford2021learning} and its derived models. A straightforward method is to crop out the image regions corresponding to each change mask and extract their global \texttt{[CLS]} tokens using CLIP, and then compute the similarity between these global \texttt{[CLS]} tokens and the text embedding of the category of interest to obtain the final identification result. However, this method requires feeding the image patches corresponding to each change mask into CLIP's image encoder, resulting in high computational cost. Inspired by SegEarth-OV~\cite{li2024segearth}, which demonstrates that patch tokens, \ie, feature map, generated by CLIP can also be used for collaborative inference with text embeddings in remote sensing scenarios. In our implementation, we extract the full-image features only once for $x_{{img}_1}$ and $x_{{img}_2}$, and then use change masks to crop the corresponding regions in the feature map, and finally calculate the average vector as the image representation of the masks (\textit{a.k.a.} masked average pooling), which is used in conjunction with the text embedding for inference.

\subsection{\IMC}

The \IMC~framework is inspired by the PCC method. It initially identify target instances of interest and determine their positions in the form of boxes, points, or masks. Subsequently, it converts these instances into a uniform mask format. Finally, the states of the bi-temporal targets at the corresponding positions are compared to determine whether any changes have occurred. Thus, the \IMC~framework can be composed of three sequential components: \textcolor{ovcdblue}{\textbf{Identifier}}, \textcolor{ovcdred}{\textbf{Mask Proposal}} and \textcolor{ovcdgreen}{\textbf{Comparator}}. 
% They are functionally different from the components in \MCI~framework.

\noindent
\textcolor{ovcdblue}{\textbf{Identifier}} here is different from that in \MCI, as it must not only identify arbitrary categories of objects but also initially discover all targets. To achieve this, some open-vocabulary detection, visual grounding, or even multi-modal large language models (MLLMs) can be selected as identifiers. For example, we process the bi-temporal images separately, using Grounding~DINOs~\cite{liu2025grounding, ren2024grounding, zhao2024open} to generate bounding boxes, Molmo~\cite{deitke2024molmo} to generate points, or APE~\cite{shen2024aligning} to generate masks directly for objects of interest.

\begin{table*}
  \caption{OVCD quantitative comparison on building change detection datasets. ``-'' denotes data missing.}
  \label{table_building}
  \centering
  \scalebox{0.85}{
  \begin{tabular}{@{}l|cc|cc|cc|cc@{}}
    \toprule[1pt]
    \multirow{2}{*}{Method} & \multicolumn{2}{c|}{LEVIR-CD} & \multicolumn{2}{c}{WHU-CD} & \multicolumn{2}{c}{S2Looking} & \multicolumn{2}{c}{BANDON} \\
    & $IoU^c$ & $F_1^c$ & $IoU^c$ & $F_1^c$ & $IoU^c$ & $F_1^c$ & $IoU^c$ & $F_1^c$\\ 
    \midrule
    PCA-KM~\cite{celik2009unsupervised} & 4.8 & 9.1 & 5.4 & 10.2 & - & - & - & - \\
    CNN-CD~\cite{el2016convolutional} & 7.0 & 13.1 & 4.9 & 9.4 & - & - & - & - \\
    DSFA~\cite{du2019unsupervised} & 4.3 & 8.2 & 4.1 & 7.8 & - & - & - & - \\
    DCVA~\cite{saha2019unsupervised} & 7.6 & 14.1 & 10.9 & 19.6 & - & - & - & - \\
    GMCD~\cite{tang2021unsupervised} & 6.1 & 11.6 & 10.9 & 19.7 & - & - & - & - \\
    CVA~\cite{bovolo2006theoretical} & - & 12.2 & - & - & - & 5.8 & - & - \\
    DINOv2+CVA~\cite{zheng2024segment} & - & 17.3 & - & - & - & 4.3 & - & - \\
    AnyChange-H~\cite{zheng2024segment} & - & 23.0 & - & - & - & 6.4 & - & - \\
    SCM~\cite{tan2023segment} & 18.8 & 31.7 & 18.6 & 31.3 & - & - & - & - \\
    \midrule
    \multicolumn{8}{@{}l}{\MCI:}\\
    \textcolor{ovcdred}{SAM}{ - }\textcolor{ovcdgreen}{DINO}{ - }\textcolor{ovcdblue}{SegEarth-OV} & 33.0 & 49.7 & 36.7 & 53.7 & 22.5 & 36.7 & 15.3 & 26.5 \\
    \textcolor{ovcdred}{SAM}{ - }\textcolor{ovcdgreen}{DINOv2}{ - }\textcolor{ovcdblue}{SegEarth-OV} & 36.6 & 53.6 & 40.6 & 57.7 & 23.9 & 38.5 & 17.6 & 30.2 \\
    \textcolor{ovcdred}{SAM2}{ - }\textcolor{ovcdgreen}{DINOv2}{ - }\textcolor{ovcdblue}{SegEarth-OV} & 33.8 & 50.5 & 40.9 & 58.1 & 23.1 & 37.6 & 17.7 & 30.1 \\
    % \textcolor{ovcdred}{SAM}{ - }\textcolor{ovcdgreen}{DINO}{ - }\textcolor{ovcdblue}{SegEarth-OV} &  &  &  &  &  &  &  & \\
    \midrule
    \multicolumn{8}{@{}l}{\IMC:} \\
    % \textcolor{ovcdblue}{Grounding DINO}{ - }\textcolor{ovcdred}{SAM2}{ - }\textcolor{ovcdgreen}{DINOv2} &  &  &  &  &  &  &  & \\
    \textcolor{ovcdblue}{MM-Grounding-DINO}{ - }\textcolor{ovcdred}{SAM2}{ - }\textcolor{ovcdgreen}{DINO} & 15.6 & 27.0 & 11.0 & 19.8 & 2.3 & 4.5 & 1.9 & 3.8 \\
    \textcolor{ovcdblue}{APE}{ - }\textcolor{ovcdred}{/}{ - }\textcolor{ovcdgreen}{DINO} & 53.5 & 69.7 & 56.8 & 72.5 & 10.1 & 18.4 & 7.8 & 14.5 \\
    \textcolor{ovcdblue}{APE}{ - }\textcolor{ovcdred}{/}{ - }\textcolor{ovcdgreen}{DINOv2} & 50.0 & 66.7 & 61.1 & 75.8 & 5.3 & 10.1 & 11.8 & 21.1 \\
    \textcolor{ovcdblue}{Grounding DINO 1.5}{ - }\textcolor{ovcdred}{SAM2}{ - }\textcolor{ovcdgreen}{DINOv2} & - & - & - & - & - & - & - & - \\
    \midrule
    \bottomrule[1pt]
  \end{tabular}}
\end{table*}

\noindent
\textcolor{ovcdred}{\textbf{Mask Proposal}} in the \IMC~framework is used to convert the bounding box, point, or coarse mask generated by the identifier into a uniform finer mask, as the OVCD task ultimately requires pixel-level output. The demand of this component directs to interactive segmentation models \eg SAM~\cite{kirillov2023segment}, SAM~2~\cite{ravi2024sam}, FastSAM~\cite{zhao2023fast}, \etc. In our implementation, we feed the bi-temporal images along with their respective identification results into the interactive segmentation model. Thus, all instance masks and their categories for each of the bi-temporal images are obtained.

% , which can utilize bounding box, point, or coarse mask as prompts to yield fine-grained target masks

\noindent
\textcolor{ovcdgreen}{\textbf{Comparator}} in the \IMC~framework has a similar capability as the comparator in the \MCI~framework, \ie, to determine whether the candidate mask region has changed. However, unlike the former, in the \IMC~framework, the categories of the candidate masks are known. Therefore, a simple geometry-based comparison method is logically sufficient. Following~\cite{li2024semicd}, we use an IoU-aware method. Specifically, an instance mask is considered as unchanged if the sum of its IoUs with all the remaining instance masks of the same category is higher than a predefined threshold; otherwise, it is regarded as changed. However, in practice, due to the limited capability of the identifier, it is common that for the identical object, it can be detected in one image but missed in another, resulting in pseudo change. To alleviate this issue, we additionally use the latent matching method in \MCI, where each masked region is individually compared at the feature level. Finally, only the change masks confirmed by both comparison methods are kept.

% This can also be seen as an accumulation of errors.

\begin{table*}
  \caption{OVCD quantitative comparison on SECOND dataset. ``-'' denotes that the score is close to 0.}
  \label{table_scd}
  \centering
  \scalebox{0.85}{
  \begin{tabular}{@{}l|cc|cc|cc|cc|cc|cc@{}}
    \toprule[1pt]
    \multirow{2}{*}{Method} & \multicolumn{2}{c|}{Building} & \multicolumn{2}{c|}{Tree} & \multicolumn{2}{c|}{Water} & \multicolumn{2}{c|}{Low vegetation} & \multicolumn{2}{c|}{N.v.g surface} & \multicolumn{2}{c}{Playground}\\
    & $IoU^c$ & $F_1^c$ & $IoU^c$ & $F_1^c$ & $IoU^c$ & $F_1^c$ & $IoU^c$ & $F_1^c$ & $IoU^c$ & $F_1^c$ & $IoU^c$ & $F_1^c$\\ 
    \midrule
    \multicolumn{8}{@{}l}{\MCI:}\\
    \textcolor{ovcdred}{SAM}{ - }\textcolor{ovcdgreen}{DINO}{ - }\textcolor{ovcdblue}{SegEarth-OV} & 34.1 & 50.8 & 16.5 & 28.3 & 13.4 & 23.6 & 24.0 & 38.7 & 22.5 & 36.7 & 16.0 & 27.6 \\
    \textcolor{ovcdred}{SAM}{ - }\textcolor{ovcdgreen}{DINOv2}{ - }\textcolor{ovcdblue}{SegEarth-OV} & 38.1 & 55.2 & 20.3 & 33.8 & 14.3 & 25.1 & 24.1 & 38.9 & 26.2 & 41.6 & 20.0 & 33.3 \\
    \textcolor{ovcdred}{SAM2}{ - }\textcolor{ovcdgreen}{DINOv2}{ - }\textcolor{ovcdblue}{SegEarth-OV} & 36.6 & 53.5 & 18.2 & 30.8 & 13.8 & 24.3 & 22.1 & 36.2 & 19.2 & 32.3 & 17.1 & 29.2 \\
    % \textcolor{ovcdred}{SAM}{ - }\textcolor{ovcdgreen}{DINO}{ - }\textcolor{ovcdblue}{SegEarth-OV} &  &  &  &  &  &  &  & \\
    \midrule
    \multicolumn{8}{@{}l}{\IMC:} \\
    % \textcolor{ovcdblue}{Grounding DINO}{ - }\textcolor{ovcdred}{SAM2}{ - }\textcolor{ovcdgreen}{DINOv2} &  &  &  &  &  &  &  &  &  &  &  & \\
    \textcolor{ovcdblue}{MM-Grounding-DINO}{ - }\textcolor{ovcdred}{SAM2}{ - }\textcolor{ovcdgreen}{DINO} & 9.5 & 17.4 & 7.0 & 13.1 & 1.2 & 2.3 & 5.2 & 9.8 & 1.0 & 2.0 & - & - \\
    \textcolor{ovcdblue}{APE}{ - }\textcolor{ovcdred}{/}{ - }\textcolor{ovcdgreen}{DINO} & 26.5 & 42.0 & 13.5 & 23.8 & 9.8 & 17.9 & - & - & - & - & 16.5 & 28.3 \\
    \textcolor{ovcdblue}{APE}{ - }\textcolor{ovcdred}{/}{ - }\textcolor{ovcdgreen}{DINOv2} & 28.1 & 43.9 & 14.1 & 24.8 & 12.2 & 21.7 & 1.4 & 2.7 & - & - & 16.0 & 27.6 \\
    % \textcolor{ovcdblue}{DINO-X}{ - }\textcolor{ovcdred}{SAM2}{ - }\textcolor{ovcdgreen}{DINOv2} &  &  &  &  &  &  &  &  &  &  &  & \\
    \bottomrule[1pt]
  \end{tabular}}
\end{table*}

\begin{table*}
  \caption{Comparison of OVCD method (unsupervised) and supervised change detection method on cross-datasets (unseen data). \textcolor{ovcdgray}{Gray} indicates training and evaluation on the same dataset, which can be seen as the upper bound of performance and is not involved in the comparison.}
  \label{cross_dataset}
  \centering
  \scalebox{0.9}{
  \begin{tabular}{@{}l|cc|cc|cc|cc@{}}
    \toprule[1pt]
    \diagbox{Train on:}{Test on:} & \multicolumn{2}{c|}{LEVIR-CD} & \multicolumn{2}{c|}{WHU-CD} & \multicolumn{2}{c|}{S2Looking} & \multicolumn{2}{c}{BANDON} \\
    & $IoU^c$ & $F_1^c$ & $IoU^c$ & $F_1^c$ & $IoU^c$ & $F_1^c$ & $IoU^c$ & $F_1^c$ \\ 
    \midrule
    LEVIR-CD & \textcolor{ovcdgray}{84.9} & \textcolor{ovcdgray}{91.8} & 50.3 & 66.9 & 1.5 & 2.9 & 3.0 & 5.8 \\
    WHU-CD & 16.9 & 28.9 & \textcolor{ovcdgray}{87.9} & \textcolor{ovcdgray}{93.5} & 3.1 & 6.0 & 3.7 & 7.1 \\
    S2Looking & 45.6 & 62.7 & 22.0 & 36.1 & \textcolor{ovcdgray}{49.6} & \textcolor{ovcdgray}{66.3} & 3.8 & 7.4 \\
    BANDON & 38.8 & 55.9 & 46.7 & 63.7 & 12.0 & 21.4 & \textcolor{ovcdgray}{52.1} & \textcolor{ovcdgray}{68.5} \\
    \midrule
    OVCD & 53.5 & 69.7 & 61.1 & 75.8 & 23.9 & 38.5 & 17.6 & 30.2 \\
    $\Delta$ & \textbf{\textcolor{tabgreen}{$\uparrow$7.9}} & \textbf{\textcolor{tabgreen}{$\uparrow$7.0}} & \textbf{\textcolor{tabgreen}{$\uparrow$10.8}} & \textbf{\textcolor{tabgreen}{$\uparrow$8.9}} & \textbf{\textcolor{tabgreen}{$\uparrow$11.9}} & \textbf{\textcolor{tabgreen}{$\uparrow$17.1}} & \textbf{\textcolor{tabgreen}{$\uparrow$13.8}} & \textbf{\textcolor{tabgreen}{$\uparrow$22.8}} \\
    \bottomrule[1pt]
  \end{tabular}}
\end{table*}

\section{Experiment}

\subsection{Dataset}

To fully assess the proposed \MCI~and \IMC~frameworks, four building change detection and one land cover change detection datasets are selected. Since both proposed frameworks are training-free, we mainly focus on their test/validation sets.
% (more data evaluations will be released in future versions of the paper)

\noindent
\textbf{LEVIR-CD dataset}~\cite{chen2020spatial} is collected from Google Earth with a spatial resolution of 0.5m/pixels. Its test set contains 128 pairs of 1,024$\times$1,024 images. When processing with the \IMC~framework, considering the small target issue in remote sensing images~\cite{li2024segearth}, we crop the original images into non-overlapping 256$\times$256 image patches.

\noindent
\textbf{WHU-CD dataset}~\cite{ji2018fully} contains 7,434 aerial image pairs with a size of 256$\times$256 and 744 for testing. Its spatial resolution is 0.2m/pixels.

\noindent
\textbf{S2Looking dataset}~\cite{shen2021s2looking} contains 1,000 1,024$\times$1,024 test data with spatial resolution of 0.5$\sim$0.8m/pixel. It is also pre-cropped into 256$\times$256 patches for \IMC~framework.

\noindent
\textbf{BANDON dataset}~\cite{pang2023detecting} consists of 392 test image pairs with a size of 2,048$\times$2,048, which are mainly collected from Google Earth, Microsoft Virtual Earth and ArcGIS. Its test set consists of 207 in-domain pairs and 185 out-domain pairs; here we use only the in-domain images. These images are pre-cropped to 1,024$\times$1,024 patches for \MCI~framework, and 256$\times$256 patches for \IMC~framework.

\noindent
\textbf{SECOND dataset}~\cite{yang2021asymmetric} focus on 6 main land-cover classes, \ie, non-vegetated ground surface (N.v.g surface), tree, low vegetation, water, building, and playground. It contains 4,662 pairs of aerial images and 593 for testing.

\subsection{Implementation Details}

\noindent
\textbf{Codebase.} Since this paper presents the OVCD task and framework for the first time, we develop a PyTorch-based codebase, \textit{Dynamic\textcolor{tabblue}{Earth}}, for the reimplementation of the proposed methods and subsequent research. The core code of \textit{Dynamic\textcolor{tabblue}{Earth}} is organized to match both frameworks, \ie, it includes identifiers, comparators, and mask proposals. In addition, \textit{Dynamic\textcolor{tabblue}{Earth}} includes both evaluation scripts for each dataset and demo scripts for each method with detailed comments and standardized code. The different scripts are isolated, and we believe this design is user-friendly to understand, debug, and add code. 

\noindent
\textbf{Setup.} All methods proposed in this paper can be run on a single 4090 GPU. The setting of the hyperparameter $\beta$ fluctuates across methods or categories, and we endeavor to find its optimal value. For text prompts, due to different preferences in various VLMs, we will discuss this below.

\noindent
\textbf{Evaluation.} For all datasets, we calculate the IoU and F1 score corresponding to each class separately, which are denoted as $IoU^c$ and $F_1^c$.

\noindent
\textbf{Compared method.} Since both frameworks proposed in this paper are training-free, we select some unsupervised methods for comparison, including PCA-KM~\cite{celik2009unsupervised}, DSFA~\cite{du2019unsupervised}, DCVA~\cite{saha2019unsupervised}, GMCD~\cite{tang2021unsupervised} and CVA~\cite{bovolo2006theoretical}. In addition, AnyChange~\cite{zheng2024segment} is used for comparison, considering single-class change detection as binary change detection. The closest to our task is SCM~\cite{tan2023segment}, which is designed for unsupervised building change detection.

%-------------------------------------------------------------------------

\begin{figure*}[t]
  \centering
   \includegraphics[width=0.98\linewidth]{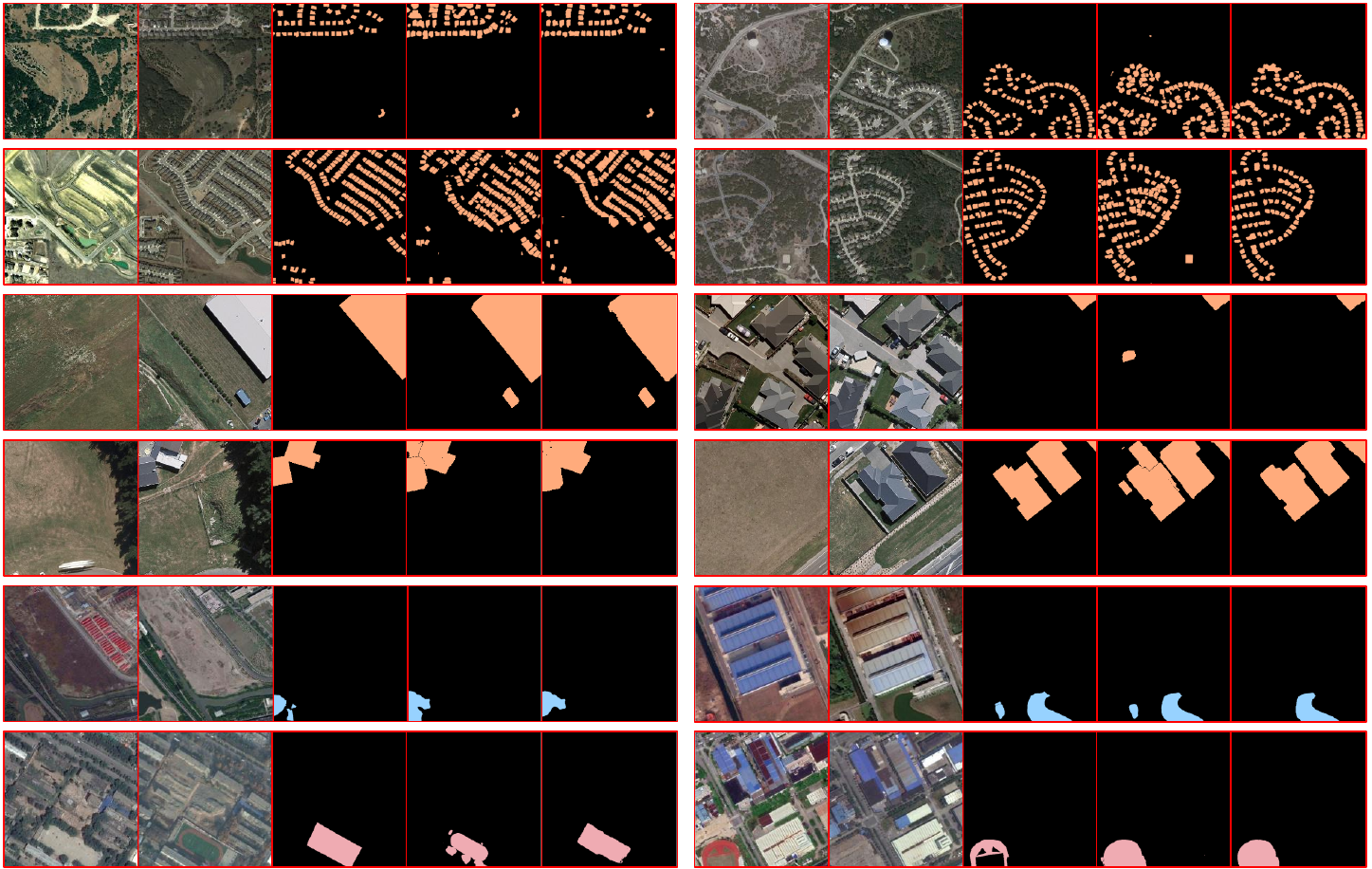}
   \caption{Open-vocabulary change detection examples. In each group: $x_{{img}_1}$, $x_{{img}_2}$, ground truth, the result of an \MCI~method and the result of an \IMC~method. Color rendering: \textcolor{ovcd_building}{\textit{``Building''}}, \textcolor{ovcd_water}{\textit{``Water''}}, \textcolor{ovcd_playground}{\textit{``Playground''}}.}
   \label{fig:res}
\end{figure*}

\subsection{Results}

\noindent
\textbf{Building Change Detection.} As listed in Table \ref{table_building}, we build several OVCD methods under the \MCI~and \IMC~frameworks. For the \MCI~framework, we use SAM or SAM~2 as the mask proposal, DINO or DINOv2 as the comparator, and SegEarth-OV as the identifier. Their performance differences are not significant, which suggests that the \MCI~framework is relatively stable, and this stability stems from the strong generalization capabilities of SAMs, DINOs and CLIPs (on which SegEarth-OV relies). The performance of the \IMC~methods is highly dependent on the identifier. According to their reports~\cite{zhao2024open, shen2024aligning}, both MM-Grounding-DINO and APE are trained only on some public general image data, but the latter has significantly superior ability in remote sensing scenarios than the former, leading to the performance difference in the OVCD task. Compared to the S2Looking and BANDON datasets, LEVIR-CD and WHU-CD are simpler and they have higher image quality and clearer context. An interesting observation is that in simple data, the methods under the \IMC~framework achieve the best performance, while in complex data, the \MCI~methods achieve the best results. This suggests that the \MCI~framework is more robust to some extent, and in some complex scenarios, suboptimal identifiers in the \IMC~framework result in serious error accumulation.

\noindent
\textbf{Land cover change detection.} Beyond buildings, there are several other land cover types that are of interest to users. In Table~\ref{table_scd}, we evaluate the OVCD of 6 categories in the SECOND dataset. It can be observed that the \MCI~methods generally outperform the \IMC~methods due to the fact that the identifiers in \IMC~can hardly recognize some categories, which are often regarded as \textit{``background''} in natural images, \eg, \textit{``Tree''}. In addition, some categories, \eg, \textit{``Low vegetation''}, \textit{``N.v.g. surface''}, are difficult to instantiate. All methods yield superior results for \textit{``Building''} compared to other categories. We suppose that this is because buildings, as one of the most common man-made land covers, have stronger \textit{``foreground''} characteristics and a certain data bias in the training data of the foundation models.

\noindent
\textbf{Effectiveness of OVCD.} The significance of OVCD is to bridge the gap between vision and language, improve the generalization of the model, and enable the model to have the ability to detect any change of interest, thus avoiding costly re-training. Therefore, in Table~\ref{cross_dataset}, we compare the cross-dataset performance of supervised learning models with the OVCD method. We select the state-of-the-art supervised change detection model, Changer~\cite{fang2023changer, li2024open}. It can be observed that the models trained on LEVIR-CD and WHU-CD are nearly unusable on S2Looking and BANDON. The models trained on S2Looking and BANDON perform reasonably well on other datasets, but are still far from the results trained on their own datasets. The OVCD method is significantly superior to the best cross-dataset results on all datasets, which confirms the availability and potential of OVCD in real-world scenarios.

\noindent
\textbf{Visualization.} In Figure~\ref{fig:res}, we visualize some OVCD results for the LEVIR-CD, WHU-CD and SECOND datasets. It can be observed that \MCI~method can handle small targets well without pre-cropping, but there are some compact targets that still cannot be instantiated and detected. The \IMC~method is highly dependent on the capability of the identifier, which ensures high precision, but some targets may be dropped at the initial stage (low recall). In addition, the proposed OVCD method is insufficient in detecting the internal components of the target (\ie, part segmentation), \eg, partial changes in the interior of the playground in the last group of samples.

\noindent
\textbf{Multi-class inference.} In the current version implementation of the proposed methods, although arbitrary category inference can be realized, it needs to be processed one by one when performing multi-class change detection. We find that fine-grained category division may impair the performance of VLM when processing remote sensing images. We believe that this is caused by the gap between the generic image domain and the remote sensing image domain. In addition, in multi-class inference, some judgment logic will be more complicated, \eg, category information may need to be considered in the comparator.

\noindent
\textbf{Design of prompts.} In inference, VLMs are sensitive to textual prompt inputs. Consider a binary segmentation scene where we generally use the template with the category name as the foreground prompt and the one with \textit{``background''} as the background prompt. Indeed, this works for the common category of change detection, \ie \textit{``Building”}. However, for the various categories in the land cover, such simple prompts may not work well. We can alleviate this issue by adding synonyms to the foreground prompt and other categories involved in the dataset to the background prompt. However, carefully designing prompt words for each category is not what we expect. Possible solutions to improve this issue are to design a comprehensive vocabulary dictionary or to use LLM to design prompts~\cite{wysoczanska2024study}.

\section{Conclusion}

In this paper, we propose a new task, open-vocabulary change detection, which realizes the connection between vision and language in change detection and can segment any change of interest. Inspired by the \textit{``4W''} assertion, we propose two frameworks for OVCD, \MCI~and \IMC. Based on these two frameworks, we instantiate several training-free OVCD methods equipped with off-the-shelf VLMs. Through comprehensive experiments on multiple change detection datasets, we show the superiority of the proposed OVCD method over previous unsupervised methods and further demonstrate the effectiveness of OVCD in practice. In addition, we contribute the first OVCD codebase, \textit{Dynamic\textcolor{tabblue}{Earth}}, to the Earth vision community for algorithm development, evaluation, and application. Although the proposed OVCD method still falls short of purely supervised methods, we believe that open-world perception is what is needed for practical change detection. We hope that subsequent research will further improve the OVCD method, either training-based or training-free, either in accuracy or efficiency, \etc.

%------------------------------------------------------------------------

{\small
\bibliographystyle{ieee_fullname}
\bibliography{egbib}
}

\end{document}